\documentclass[10pt,twocolumn,letterpaper]{article}

\usepackage{cvpr}
\usepackage{times}
\usepackage{epsfig}
\usepackage{graphicx}
\usepackage{amsmath}
\usepackage{amssymb}

\usepackage{multirow}
\usepackage{bigstrut}
\usepackage{booktabs}

\usepackage[breaklinks=true,bookmarks=false]{hyperref}
\cvprfinalcopy % *** Uncomment this line for the final submission

\def\cvprPaperID{****} % *** Enter the CVPR Paper ID here

\ifcvprfinal\fi
\begin{document}

%%%%%%%%% TITLE
\title{Excavate Condition-invariant Space by Intrinsic Encoder}

\author{Jian Xu,
Chunheng Wang,
Cunzhao Shi,
and
Baihua Xiao\\
Institute of Automation, Chinese Academy of Sciences (CASIA)\\
%\and
}

\maketitle
%\thispagestyle{empty}

%%%%%%%%% ABSTRACT
\begin{abstract}
   %人的认知是在语义层面的，树木，山，路等，例如可以不受树木是冬天的秃树还是夏天的绿树的影响。
    As the human, we can recognize the places across a wide range of changing environmental conditions such as those caused by weathers, seasons, and day-night cycles. We excavate and memorize the stable semantic structure of different places and scenes. For example, we can recognize tree whether the bare tree in winter or lush tree in summer. Therefore, the intrinsic features that are corresponding to specific semantic contents and condition-invariant of appearance changes can be employed to improve the performance of long-term place recognition significantly.

    In this paper, we propose a novel intrinsic encoder that excavates the condition-invariant latent space of different places under drastic appearance changes. Our method excavates the space of intrinsic structure and semantic information by proposed self-supervised encoder loss. Different from previous learning based place recognition methods that need paired training data of each place with appearance changes, we employ the weakly-supervised strategy to utilize unpaired set-based training data of different environmental conditions.

    We conduct comprehensive experiments and show that our semi-supervised intrinsic encoder achieves remarkable performance for place recognition under drastic appearance changes. The proposed intrinsic encoder outperforms the state-of-the-art image-level place recognition methods on standard benchmark Nordland.
\end{abstract}

\section{Introduction}
\begin{figure}
  \centering
  \includegraphics[width=3 in]{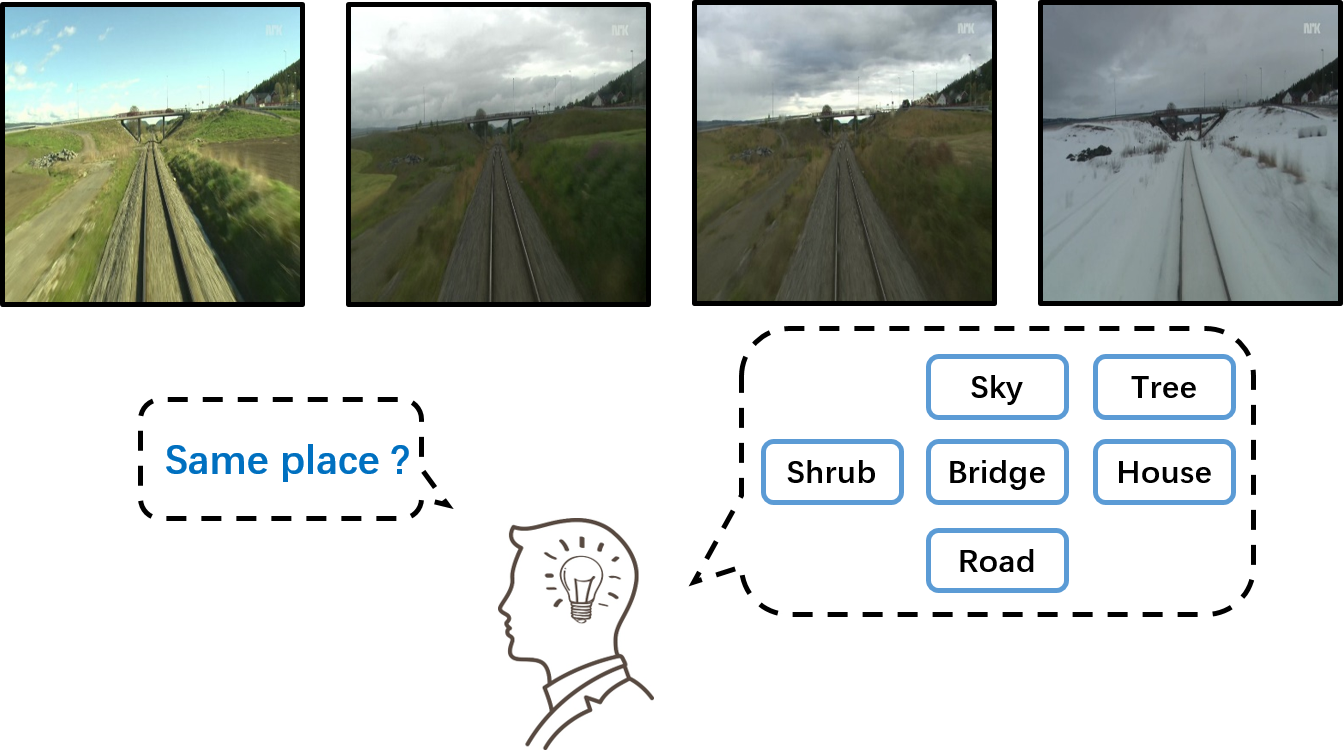}\\
  \caption{
  Human have strong cognitive competence based on semantic comprehension and environmental adaptation.
  As the human, we can capture the intrinsic and semantic information under drastic appearance changes.
  }
  \label{image_show}
\end{figure}

%人可以轻松的辨别出long-term的同一个场景，例如无论是春夏秋冬，如图所示，人可以抓住关键信息，无论什么季节，什么天气，人可以辨认出树木，桥梁，铁路，围栏等物体，并且根据位置判断是不是同一个场景。人的较强的认知能力基于语义理解能力和对于环境的适应性，人可以不被外表干扰而捕获到本质的语义信息。
People can easily recognize places under environmental disturbance. For example, when we see the places in various seasons, weather, and illumination as shown in Fig.~\ref{image_show}, we can deduce that four images are in the same place. Why can human solve place recognition even under drastic appearance changes$?$ One reason is that human can distinguish trees, bridge, road, house under environmental disturbance. Human have strong cognitive competence based on semantic comprehension and environmental adaptation, we can capture the intrinsic and semantic information under drastic appearance changes.
The challenging task, given an image of a place, can a human, animal, or robot decide whether or not this image is of a place it has already seen, is defined as visual place recognition~\cite{VPL_Survey}.
%从人的认知引入，然后介绍应用任务place recognition.
%Visual place recognition~\cite{VPL_Survey} is a well-defined but extremely challenging problem to solve in the general sense; given an image of a place, can a human, animal, or robot decide whether or not this image is of a place it has already seen$?$

In recent years, visual place recognition has received a significant amount of attention both in computer vision~\cite{Graphbased-pl,repetitive-pl,24-7-pl,netvlad,netvlad_pami,crn} and robotics
communities~\cite{FAB-MAP,night-day-pl,illumination-invariance-pl,deep-pl,gan-pl}, motivated by its wide range of applications including autonomous driving~\cite{changing-conditions-db}, adding and refining geo-tags in image collections~\cite{image-collections} and augmented reality~\cite{augmented-reality}.
The perceptual changes, caused by factors such as day-night cycles, varying weather conditions and seasonal change, still remain a significant challenge for long-term place recognition.
The key issue is how to extract the condition-invariant representation of a place under drastic appearance changes.
Recently, much attention is given to two respects: feature-level and image-level condition-invariant representation.

%特征级
Feature-level methods focus on extracting condition-invariant features under drastic appearance changes.
Traditionally hand-crafted features such as SIFT~\cite{sift} and SURF~\cite{surf} are used extensively in many computer vision tasks.
The variants of SIFT, U-SIFT~\cite{u-sift}, is shown to be robust of lighting invariance.
Instead of utilizing point-based features, other features such as edge-based features and whole image descriptors can be chosen.
Edge features~\cite{edge-pl} are invariant to lighting, orientation and scale.
SeqSLAM~\cite{seqslam} and other sequence-based and trajectory-based methods~\cite{fast-seqslam,flow-pl,robust-pl,robust-pl2,trajectory-pl} are proposed to do localization across different environmental conditions.
After that, the methods~\cite{pre-cnn-pl} based on Convolutional Neural Network (CNN)~\cite{cnn} are introduced to solve the environmental changes.
Pre-trained CNN are utilized to extract viewpoint- and condition-invariant features~\cite{pre-cnn-pl}.
Some recent studies~\cite{cnn-pl,deep-pl,netvlad,netvlad_pami,crn} train the CNN for place recognition based on collected training datasets by triplets loss or classification loss.
However, these learning-based methods need collect paired training data which consist of the paired images in the same position and under different environmental conditions.
In practice, the collection of paired training data is laborious and time-consuming.
The domain-based set-wise training data are relatively easier to collect.

%图像级
Image-level methods pay attention to capturing intrinsic scene appearance under various conditions.
In order to reduce the disturbance of environmental changes, some works~\cite{transform_mn,superpixel_appearance_change,gan-pl} transform the image from original domain (e.g., morning and winter) to other domain (e.g., afternoon and summer) as the image preprocessing for place recognition.
However, we have to know which domain the image is in before the translation of domains.
By contrast, the methods~\cite{deal_shadows,illumination-invariance-pl,physics_shadow_invariant} which capture intrinsic scene appearance under various conditions by removing environmental changes without knowing the domain of the image.
These removal methods are based on hand-crafted models and empirical analysis, and each method can only solve one change.

Inspired by strong cognitive competence of human vision, we propose an intrinsic encoder that captures condition-invariant intrinsic structure and semantic information in computer vision to simulate human cognition in this paper.
%解释上可以强调语义
The proposed intrinsic encoder is trained based on set-wise images of various seasons, weathers and day-night cycles to excavate condition-invariant representation space, which can work as a image preprocessor for long-term place recognition.
Because pixel-wise corresponding training data are hard to collect in practice, we design a self-supervised encoder loss function to train intrinsic encoder by weakly-supervised unpaired training data rather than paired pixel-wise corresponding data motivated by CycleGAN~\cite{CycleGAN}.
Our intrinsic encoder can capture the intrinsic feature space corresponding to specific semantic contents implicitly and adapt to various environmental changes.
As far as we know, it is the first method that can handle various environmental changes simultaneously without paired training data.

The main contributions of this paper can be summarized as follows:
\begin{itemize}
\item
Various natural variation in appearance caused by such things as weather phenomena, diurnal fluctuation in lighting, and seasonal changes can be removed by our intrinsic encoder.
One intrinsic encoder can solve multiple different environmental changes simultaneously.

\item
Because paired training data which are pixel-wise corresponding are difficult to collect in practice, we proposed self-supervised encoder loss to optimize intrinsic encoder.
Therefore, our intrinsic encoder can be trained based on weakly-supervised set-wise data rather than paired training data.

\end{itemize}

\section{Related Work}
\subsection{Long-term Place Recognition}
For visual place recognition task under severe appearance changes, the methods can be broadly divided into two following dominant branches:
$(1)$ feature-level methods that represent an image as a set of condition-invariant features descriptors extracted from interesting locations in the image, and $(2)$ image-level methods that try to capture intrinsic scene appearance under various conditions.

Feature-level methods perform well in the short time frame where changes in illumination and appearance are limited.
The hand-crafted U-sift~\cite{u-sift} has greater illumination invariance among traditional point-based features~\cite{SIFT_SURF_seasons}.
Instead of point-based features, the whole image descriptors are utilized in SeqSLAM~\cite{seqslam} to solve place recognition under extreme weather changes.
Convolutional Neural Network (CNN) is successfully employed in place recognition due to their invariance of illumination and viewpoint changes~\cite{pre-cnn-pl}.
NetVLAD~\cite{netvlad,netvlad_pami} and CRN~\cite{crn} train the network containing novel aggregation layer by triplet loss for place recognition based on collected paired training data, which consist of many images under various conditions at the same place.
Additionally, the multi-scale feature encoding CNN~\cite{deep-pl} are trained to generate condition- and viewpoint-invariant features based on a massive paired Specific PlacEs Dataset (SPED) with hundreds of examples of place appearance change at thousands of different places.
However, these learning-based methods depend on the collected paired training data which consist of the paired images under different environmental conditions at the same place.

Feature-level methods fail in situations where the environmental changes are greater than their invariance of rotation, scale, illumination and so on, which are offered by the underlying feature descriptors.
Image-level methods are proposed to capture intrinsic scene appearance under extreme changes.
These approaches can be further divided into two broad categories:
$(1)$ transforming the image from original domain to another, and $(2)$ removing environmental changes.
Belonging to the first category, superpixel-based appearance change prediction~\cite{superpixel_appearance_change}, linear regression~\cite{transform_mn}, and Generative Adversarial Network (GAN)~\cite{gan-pl} are employed to address place recognition as a domain translation task.
Belonging to the second category, the intrinsic scene appearance is captured by removing shadows based on spectral response in~\cite{deal_shadows}.
Colin, et. al.,~\cite{illumination-invariance-pl} map the original image to an illumination-invariant chromaticity space to remove the appearance changes of illumination.
Recently, Relit Spectral Angle-Stacked Autoencoder (RSA-SAE)~\cite{physics_shadow_invariant} is proposed to learn an illumination-invariant mapping based on a physics-based model for illumination.
However, we need to know original domain before employing the translation of domains Additionally, the removal-based methods are based on hand-crafted models and empirical analysis, and each method can only solve one change.

\subsection{GAN}
%GAN
Generative Adversarial Networks (GAN)~\cite{GAN} achieves impressive results in image generation~\cite{LAPGAN,DCGAN}, image editing~\cite{image_edit_gan}, image inpainting~\cite{inpaint_gan}, representation learning~\cite{representation_gan,DCGAN}, and image translation~\cite{pix2pix,DiscoGAN,DualGAN,CycleGAN}.
In its simplest form, a GAN consists of two components: a generator $G$, which randomly samples from a latent space and aims to generate a realistic image that resembles images from a domain being learned, and a discriminator $D$ whose task is to correctly discriminate between real and generated (fake) images.
Recently, CycleGAN~\cite{CycleGAN} proposes the cycle consistency loss to utilize unpaired training data for image translation.
%remove domain gaps
Compared with GANPL~\cite{gan-pl} that removes domain gaps by transforming the image from original domain to target domain, we learn a robust intrinsic encoder that captures latent semantic space to bridge domain gaps in environmental changes.
%and some recent Person re-identification (re-ID) methods~\cite{reid_gan1,reid_gan2}

%VAE
Variational autoencoder (VAE)~\cite{VAE,VAE1} optimizes a variational bound.
VAE consists of an encoder and a decoder, and the prior distribution of latent variables is Gaussian typically.
Some recent works~\cite{VAE2,VAE3} generate better images by improving the variational approximation.
%VAE-GAN （VAE-GAN, BGAN, CoGAN）
The VAE-GAN ~\cite{VAE-GAN} architecture is proposed to improve image generation quality of VAE.
VAE-GAN is structured by combining a VAE with a GAN, and it consists of an encoder, a generator(decoder) and a discriminator.
Similar to VAE, the KL divergence terms in loss function penalize deviation of the distribution of the latent codes from the prior distribution in VAE-GAN.
The recent work~\cite{VAE-GAN1} employs VAE-GAN based on Coupled GANs~\cite{CoGAN} for unsupervised image-to-image translation.
BGAN~\cite{BGAN} learns the discriminative binary representations for image retrieval based on VAE-GAN architecture.
%我们采取了encoder, a generator(decoder) and a discriminator的结构，但是encoder 的输出不是约束至prior distribution， 而是通过自监督encoder loss 获得环境不变性的encoder，capture本质的稳定的特征。
In this paper, we also employ the construction of encoder, generator(decoder) and discriminator to train our intrinsic encoder.
Different with VAE-GAN, we capture the stable representations based on self-supervised encoder loss, rather than the latent space (output of our intrinsic encoder) constrained to prior distribution.
It is worth noting that we only employ one encoder rather than several encoders for various domains in previous image translation works~\cite{VAE-GAN,VAE-GAN1}.

\begin{figure*}
  \centering
  \includegraphics[width=6.5 in]{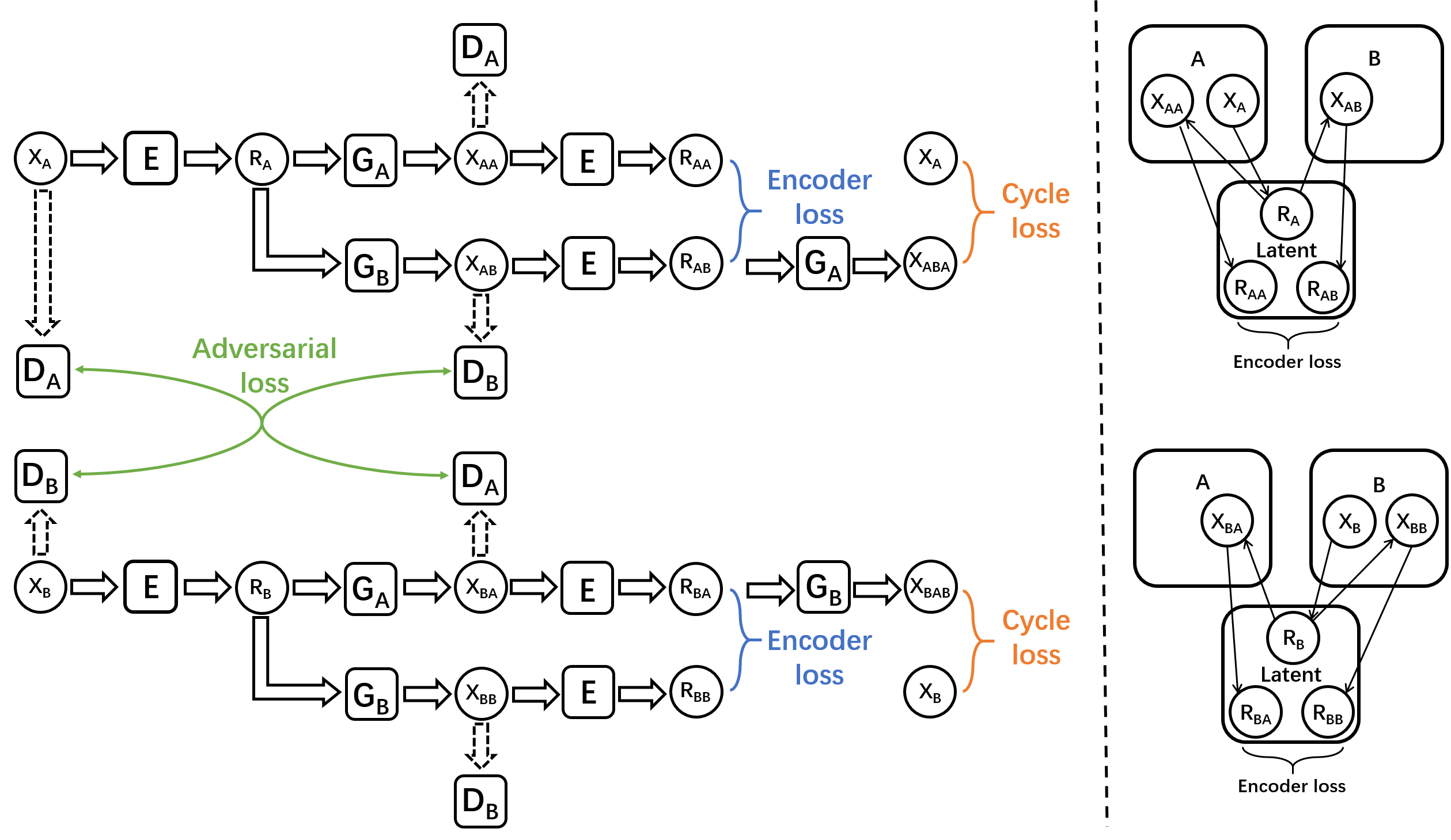}\\
  \caption{
  Architecture of our model and loss functions.
  %网络组成
  (a) Our model contains an intrinsic encoder $E$ and various generators ($G_{A}$ and $G_{B}$) and discriminators ($D_{A}$ and $D_{B}$) for different domains ($A$ and $B$).
  The encoder $E$ is the core of our network, and it can capture the stable and intrinsic representations (e.g. $R_{A}$, $R_{B}$ and $R_{AB}$) from input images (e.g. $X_{A}$, $X_{B}$ and $X_{AB}$ respectively).
  $D_{A}$ encourages $G_{A}$ to generate outputs (e.g. $X_{AA}$, $X_{BA}$ and $X_{ABA}$) indistinguishable from domain $A$, and vice versa for $D_{B}$ encourages $G_{B}$.
  %损失函数
  Adversarial loss~\cite{GAN} forces the distribution of generated images to the data distribution in the original domains.
  Cycle loss~\cite{CycleGAN} encourages that an image translated from one domain to other and retranslated back should match the original image.
  To further regularize the encoder, we introduce self-supervised encoder loss that captures the intuition that if we encode the input images from different domains in the same place we should get same intrinsic representations.
  Note that the paired images from different domains in the same place are generated based on our encoder and generators rather than paired real training images.
  %图像域
  (b) The real training images ($X_{A}$ and $X_{B}$), generated images ($X_{AA}$, $X_{AB}$, $X_{BB}$, $X_{BA}$) and representations (($R_{A}$, $R_{B}$), $R_{AA}$, $R_{AB}$, $R_{BB}$, $R_{BA}$) are divided into different domains ($A$, $B$ and $Latent$ $space$).
  }
  \label{architecture}
\end{figure*}

\section{Method}
Inspired by human visual perception, we propose an intrinsic encoder to simulate robust and power human cognition.
%符号说明
Our goal is to learn an encoder $E$, which we called intrinsic encoder in this paper, that capture stable representations ($R_{A}$ and $R_{B}$) in latent space of various domains ($A$ and $B$) given training images ($X_{A}$ and $X_{B}$).
As illustrated in Fig.~\ref{architecture}, our network contains an intrinsic encoder $E$, various generators ($G_{A}$ and $G_{B}$) and discriminators ($D_{A}$ and $D_{B}$) for different domains ($A$ and $B$).
Our objective consists of three types of terms:
(1) encoder loss to enforce encoder ($E$) to capture intrinsic information in latent space from various domains ($A$ and $B$).
(2) adversarial loss~\cite{GAN} for matching the distribution of generated images to the data distribution in the original domains ($A$ and $B$).
(3) cycle loss~\cite{CycleGAN} to encourage an image translated from one domain to other and retranslated back to match the original image and guarantee that we could recover the image from its intrinsic representation.

%网络结构
\subsection{System Architecture}
Our framework consists of an encoder ($E$), generators ($G_{A}$ and $G_{B}$) and discriminators ($D_{A}$ and $D_{B}$).
In our network, the generators and discriminators are different for various domains, while there is only one encoder which can remove environmental changes for various domains.

%每个部分的作用，输入 输出

%E
Intrinsic encoder $E$ is the core part of our network.
The inputs of it are the images in various domains, and outputs are intrinsic representations of the same size (height and width) as input images in latent space.
It captures the intrinsic and stable information of images under various conditions and domains, and it removes the interference information that changes in different domains.

%G
Generators $G_{A}$ and $G_{B}$ are different for various domains $A$ and $B$.
Generator aims to generate indistinguishable fake images based on intrinsic representations in latent space.
It generates the paired images in different domains to train the encoder by self-supervised strategy.

%D
Discriminators $D_{A}$ and $D_{B}$ are different for various domains $A$ and $B$.
Discriminator tries to distinguish between real images and generated images.
The competition between discriminator and generator forces the generated images to be indistinguishable from the real image.

\subsection{Intrinsic Encoder}
\begin{figure}
  \centering
  \includegraphics[width=3.0 in]{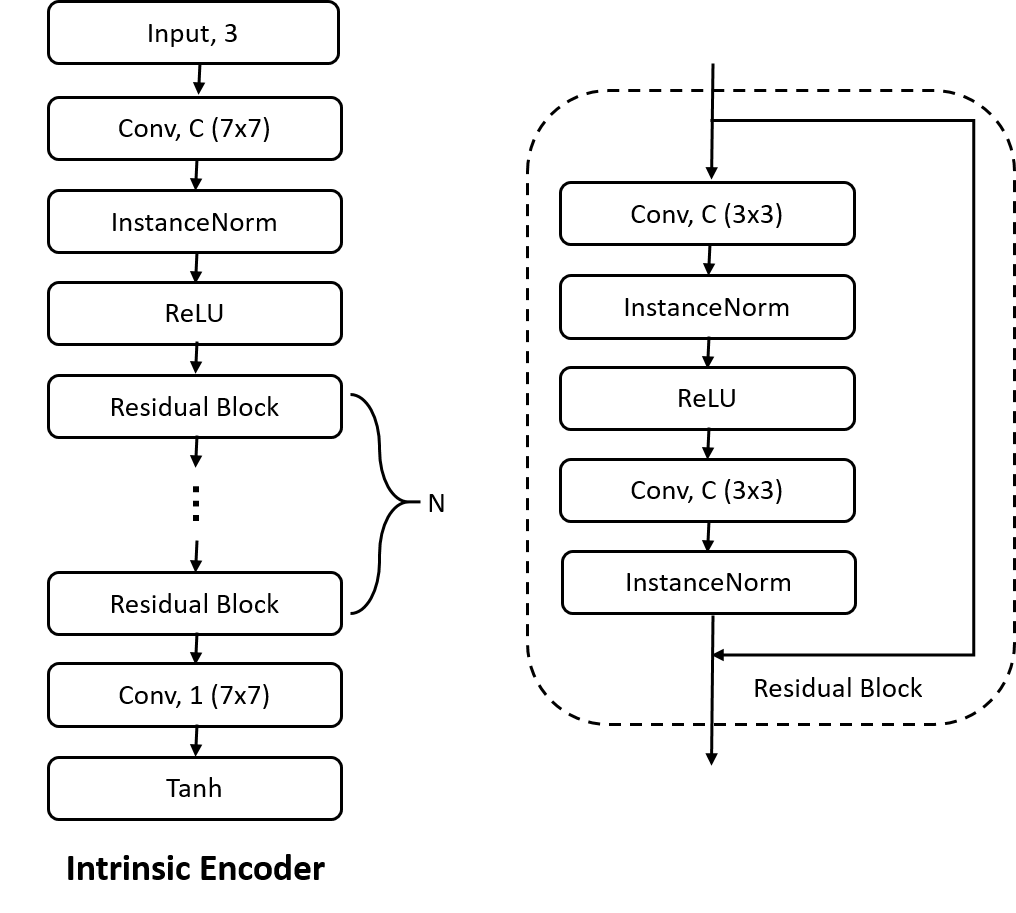}\\
  \caption{Configurations of an intrinsic encoder. The intrinsic encoder (left) consists of convolution layer, instance normalization layer, ReLU layer, residual block~\cite{resnet} and tanh layer.
  Residual block~\cite{resnet} (right) is the significant part of our intrinsic encoder, and it includes convolution layer, instance normalization layer and ReLU layer.
  }
  \label{network}
\end{figure}
As depicted in Fig.~\ref{network}, proposed intrinsic encoder consists of convolution layer, instance normalization layer, ReLU layer, residual block~\cite{resnet} and tanh layer.
In order to prevent the loss of detailed information, we do not employ pooling layer in our intrinsic encoder.
Pooling layer is generally utilized in Convolutional Neural Network (CNN) and Generative Adversarial Network (GAN) to achieve partial rotation invariance and scale invariance.
However, the down-sampling by pooling layer leads to loss of subtle information to some degree.
If down-sampling is used, deconvolution layer~\cite{deconvolution} or sub-pixel layer~\cite{sub-pixel} is generally employed to up-sampling.
In order to prevent the loss of intrinsic information, we discard the down-sampling layer and up-sampling layer in our intrinsic encoder.

We repeats residual block~\cite{resnet} several times in our intrinsic encoder, and the configuration of residual block~\cite{resnet} is depicted in Fig.~\ref{network} (right) in detail.
It consists of convolution layer, instance normalization layer and ReLU layer.
Residual block~\cite{resnet} is widely used in computer vision, and it achieves remarkable performance on many tasks.
The deep network based on residual block~\cite{resnet} is more easy to train than a plain network.
Our intrinsic encoder employs N residual blocks to extract high-level intrinsic and robust representations.

%损失函数
\subsection{Loss Function}
\subsubsection{Adversarial Loss}
In order to force the generator to generate more deceptive images, we employ adversarial loss~\cite{GAN} of generator and discriminator in our loss function.
Generator and discriminator try to defeat each other: the generator aims to produce images that the discriminator would misclassify them as realistic images, and the discriminator aims to correctly identify which image is real and which is generated.
The adversarial loss~\cite{GAN} is applied to different domains.
For domain $A$, we express the objective as:
\begin{equation}\label{1}
{L_{{\rm{adv}}}(A)} = \log ({D_A}({X_A})) + \log (1 - {D_A}({G_A}(E({X_B}))))
\end{equation}
Where representations $R_{B}=E({X_B})$ are the intrinsic representations in latent space. $G_{A}$ aims to generate images $X_{BA}=G_{A}(R_{B})$ that look similar to images from domain $A$, while $D_{A}$ tries to distinguish between generated images $X_{BA}$ and real images $X_A$.
$G_{A}$ aims to minimize this objective against an adversary $D_{A}$ that tries to maximize it, i.e. $min_{G_{A}} max_{D_{A}} L_{adv}(A)$.
For domain $B$, we introduce a similar adversarial loss as well:
\begin{equation}\label{2}
{L_{{\rm{adv}}}(B)} = \log ({D_B}({X_B})) + \log (1 - {D_B}({G_B}(E({X_A}))))
\end{equation}

\subsubsection{Cycle Loss}
In order to guarantee that we can recover the image from its intrinsic representation, we utilize cycle loss~\cite{CycleGAN} which works as a regularization term in our full loss function.
Cycle loss~\cite{CycleGAN} encourages that we should arrive at where we started if we translate from one domain to the other and back again.
Therefore, intrinsic representation contains most discriminative information from which we can recover indistinguishable image similar to the real image.
The cycle loss for domain $A$ is defined as follows:
\begin{equation}\label{3}
{L_{{\rm{cyc}}}}(A) = {\left\| {{G_A}(E({G_B}(E({X_A})))) - {X_A}} \right\|_1}
\end{equation}
Where representations $R_{A}=E({X_A})$ are the intrinsic representations of real images $X_{A}$.
Adversarial loss makes generate images $X_{AB}=G_{B}(R_{A})$ look similar to images from domain $B$.
Representations $R_{AB}=E(X_{AB})$ are intrinsic representations of generated images $X_{AB}$.
Cycle loss tries to make generate images $X_{ABA}=G_{A}(R_{AB})$ consistent with corresponding real images $X_{A}$.
For domain $B$, we also introduce a similar cycle loss ${L_{{\rm{cyc}}}}(B)$:
\begin{equation}\label{4}
{L_{{\rm{cyc}}}}(B) = {\left\| {{G_B}(E({G_A}(E({X_B})))) - {X_B}} \right\|_1}
\end{equation}

\subsubsection{Encoder Loss}
In order to restrict that the intrinsic encoder can capture the stable and intrinsic representations of images in different domains, we introduce the encoder loss which minimizes the distinction of intrinsic representations of generated paired images in different domains.
Adversarial loss~\cite{GAN} and cycle loss~\cite{CycleGAN} is the basis of self-supervised encoder loss, and they can be interpreted as the regularization term.
They try to guarantee the criterion that intrinsic representations can be recovered to images, and encourage the intrinsic representations to keep more information.
We introduce an encoder loss ${L_{{\rm{enc}}}}(A)$ for domain $A$ as follows:
\begin{equation}\label{5}
{L_{enc}}(A) = {\left\| {E({G_A}(E({X_A}))) - E({G_B}(E({X_A})))} \right\|_1}
\end{equation}
Where representations $R_{A}=E({X_A})$ are the intrinsic representations of real images $X_{A}$.
We generate paired images $X_{AA}={G_A}(E({X_A}))$ and $X_{AB}={G_B}(E({X_A})))$ in domains $A$ and $B$ respectively.
The distinction of representations of generated paired images $X_{AA}$ and $X_{AB}$ is minimized to capture intrinsic information and remove unstable information in various domains.
For domain $B$, we also introduce a symmetric encoder loss ${L_{{\rm{enc}}}}(B)$:
\begin{equation}\label{6}
{L_{enc}}(B) = {\left\| {E({G_A}(E({X_B}))) - E({G_B}(E({X_B})))} \right\|_1}
\end{equation}
Encoder loss is the key to tame various domains to intrinsic space.
In order to overcome the dependence of data, we meticulously design the encoder loss free from pixel-wise data.

%训练
\subsection{Training}
Our full objective is:
\begin{equation}\label{7}
\begin{array}{l}
{L_{full}} = {L_{{\rm{adv}}}}(A) + {L_{{\rm{adv}}}}(B)\\
\qquad \quad + \alpha{L_{cyc}}(A) + \alpha{L_{cyc}}(B)\\
\qquad \quad + \beta{L_{enc}}(A) + \beta{L_{enc}}(B)
\end{array}
\end{equation}
Where $\alpha$ and $\beta$ control the relative importance of various losses.
We aim to solve the various losses jointly as follows:
\begin{equation}\label{8}
{E^*} = \mathop {min}\limits_{E,{G_A},{G_B}} \mathop {\max }\limits_{{D_A},{D_B}} {L_{full}}
\end{equation}

Inheriting from GAN, training of proposed framework results in solving a mini-max problem where the optimization aims to find a saddle point.
We apply an alternating gradient update scheme similar to the strategy described in GAN~\cite{GAN} to solve Eq.~\ref{8}.
Specifically, we first apply a gradient ascent step to update $D_{A}$ and $D_{B}$ with fixed $E$, $G_{A}$ and $G_{A}$.
Then we update $E$, $G_{A}$ and $G_{A}$ with fixed $D_{A}$ and $D_{B}$ by gradient descent step.

\section{Experiments}
Intrinsic encoder provides an effective way to remove the long-term conditions variation.
It can be used for capturing intrinsic and stable representations under drastic appearance changes.
We extensively conduct some qualitative and quantitative experiments on Nordland database~\cite{fast-seqslam, gan-pl} and the results indicate that our intrinsic encoder achieves remarkable performance.

\subsection{Settings and Database}
\subsubsection{Database}
We demonstrate the performance of our system using the Nordland\footnote{https://nrkbeta.no/2013/01/15/nordlandsbanen-minute-by-minute-season-by-season/} database. The database has been recorded on board a train in Norway, over four seasons: spring, summer, autumn and winter. Nordland dataset also contains changes of weather (e.g. sunny, cloudy, rainy and snowy) and illumination as shown in Fig.~\ref{image_show}.
Therefore, it is particularly suitable for the evaluation of our method.
The footage consists of about 10 hours of the video under each weather condition which has been synchronized using GPS information.
We extract about 3600 images from each video for the training set.
In order to train our intrinsic encoder by unpaired set-based training data, we randomly pick unpaired training data from various domains in each step.
For the testing set, we extract about 3600 images frames with no overlap with the training set.
%For the training set, we extract image 30 minutes from the video at a frame rate of 2HZ, starting approximately at 2 hours into the drive.
%We extract image frames from 3 hours to 4 hours at 5HZ for the testing set.
All the images are resized to $100\times100$ pixels.

\subsubsection{Implementation Detail}
We employ intrinsic encoder, generator and discriminator to construct our self-supervised network.
For the generator network, we use the excellent architecture successfully used in style transfer~\cite{CycleGAN} and super-resolution~\cite{super-resolution}.
It consists of two stride-2 convolution layers, nine residual blocks~\cite{resnet} with instance normalization~\cite{instance_normalization} and ReLU layer, and two fractionally-strided convolution layers with stride $\frac{1}{2}$.
For the discriminator network, we employ the fully convolutional architecture used in ~\cite{pix2pix,CycleGAN}, which can be applied to arbitrarily-sized images.
Five convolution layers with instance normalization~\cite{instance_normalization} and  LeakyReLU layer with slope 0.2 contain 64, 128, 256, 512, and 1 kernels respectively, followed by a Sigmoid function.
As an exception to the above description, instance normalization~\cite{instance_normalization} is not applied to the first convolution layer in discriminator network.
For the intrinsic encoder network depicted in Fig.~\ref{network}, we set the numbers of residual blocks~\cite{resnet} and convolution kernels to be $N=4$ and $C=64$ respectively throughout our experiments.

The weight coefficients $\alpha$ and $\beta$ controlling the relative importance of various losses are set as $\alpha=10$ and $\beta=1$ respectively.
For network training, the learn rate is set to $lr = 2\times 10^{-5}$ and the training batch size is 1. We use the Adam solver~\cite{Adam} with parameters of $beta1 = 0.5$, $beta2 = 0.999$ and $epsilon = 10^{-8}$ to optimize the parameters of our intrinsic encoder, generator and discriminator.

%如何针对于多种环境变化，方法部分仅仅描述了两种，例如初夏秋冬四个季节。
%可以直接两两配对，我们使用了这种方法，很简单，但是经过实验上验证是有效的。当然也可以使用交叉的encoder loss。
The architecture of our model and loss functions for two domains can be easily extended to adapt more than two domains (e.g., spring, summer, autumn and winter).
We can pair off any two domains simply. Although this strategy is straightforward, it achieves remarkable performance in our experiments.
%As an alternative, we also can employ crossed encoder loss.

\subsection{Evaluation}
%为了定量分析 使用place recognition的结果。
%seqslam作为baseline
%我们进行四个季节的实验，以冬天建图，其他季节检索（因为冬天变化最大）。
%1.使用固定约束下（len=4, dis<3）的acc  2. precision-recall curve (PR 曲线)
We conduct both qualitative and quantitative experiments in this paper.
In order to analyze the results quantitatively, we apply our intrinsic encoder to long-term place recognition task~\cite{seqslam,fast-seqslam,gan-pl} which aims at localization under drastic appearance changes.
For comparison, SeqSLAM~\cite{seqslam} is employed as a baseline for long-term place recognition because our intrinsic encoder is the first method that can solve multiple different environmental changes simultaneously in place recognition as far as we know.
In our method, we use our intrinsic representations to replace original images in SeqSLAM~\cite{seqslam}.
Because appearance changes between winter and other seasons are drastic especially, we evaluate the results of place recognition across winter and other three seasons in our experiments.
We report the  place recognition accuracy with sequence length $len = 1, 4$ in~\cite{seqslam} and maximum distance $dis<=1$.
%Additionally, we also give the precision-recall curves that are commonly used
%in evaluation of place recognition~\cite{seqslam,fast-seqslam,gan-pl}.

\begin{table*}[htbp]
  \centering
  \caption{
   Accuracy comparison with the intrinsic encoder without $L_{enc}$ and intrinsic encoder trained only by Summer-Winter data.
   We report the place recognition accuracy with various sequence length $len = 1, 4$ in SeqSLAM~\cite{seqslam} and maximum distance $dis <= 1$.
   The results demonstrate that the ability of generalization of our intrinsic encoder is well and our encoder loss $L_{enc}$ is important for convergence of intrinsic encoder.
  }
    \begin{tabular}{cccccccccc}
    \toprule[1.25pt]
    \multirow{2}[0]{*}{\textbf{Methods}} & \multicolumn{2}{c}{\textbf{Spring-Winter}} & \multicolumn{2}{c}{\textbf{Summer-Winter}} & \multicolumn{2}{c}{\textbf{Autumn-Winter}} \\
    \cline{2-7}
    & 1     & 4      & 1     & 4       & 1     & 4     \\
    \toprule[1.25pt]
    Intrinsic Encoder without $L_{enc}$ & 61.25 & 87.22  & 40.00 & 75.97 & 26.67 & 57.92 \\
    Intrinsic Encoder by Summer-Winter & 72.50 & 92.64  & 55.00 & 84.03 & 37.63 & 69.03 \\
    Intrinsic Encoder & \textbf{81.67} & \textbf{94.72}  & \textbf{62.29} & \textbf{86.94}  & \textbf{44.58} & \textbf{76.37} \\
    \toprule[1.25pt]
    \end{tabular}%
  \label{ablation}%
\end{table*}%

\subsection{Results and  Discussions}
\subsubsection{Ablation Study}
We investigate the contributions of proposed encoder loss by an ablation study.
%无encoder loss损失和有损失的区别
%只用SM-W训练encoder， 测试A-SP,证明泛化能力。（对比四个数据集都用）
As shown in Table.~\ref{ablation}, we show the contribution of proposed encoder loss $L_{enc}$.
The ablation study shows that our encoder loss $L_{enc}$ is important for convergence of intrinsic encoder.
The encoder loss $L_{enc}$ forces the intrinsic encoder to capture the stable and robust representations of images in different domains.
The intrinsic representations trained based on proposed encoder loss $L_{enc}$ for the same place in different domains are similar, which contain special discriminative information of semantic contents and patterns.

In order to prove the ability of generalization of proposed intrinsic encoder, we compare intrinsic encoder trained only by Summer-Winter data with it trained based on all four seasons data in Table.~\ref{ablation}.
The intrinsic encoder trained only by Summer-Winter data still achieves remarkable performance for other changing environmental conditions.
The results demonstrate that the ability of generalization of our intrinsic encoder is well.
Therefore, we can only consider a few typical environments in training stage rather than all different weather phenomena, lighting scenario, and seasonal changes in practice.
It is very impractical and time-consuming to consider all different weather phenomena, lighting scenario, and seasonal changes.
The good generalization ability of our intrinsic encoder is significant in practice.

\subsubsection{Qualitative Results}
\begin{figure}
  \centering
  \includegraphics[width=3 in]{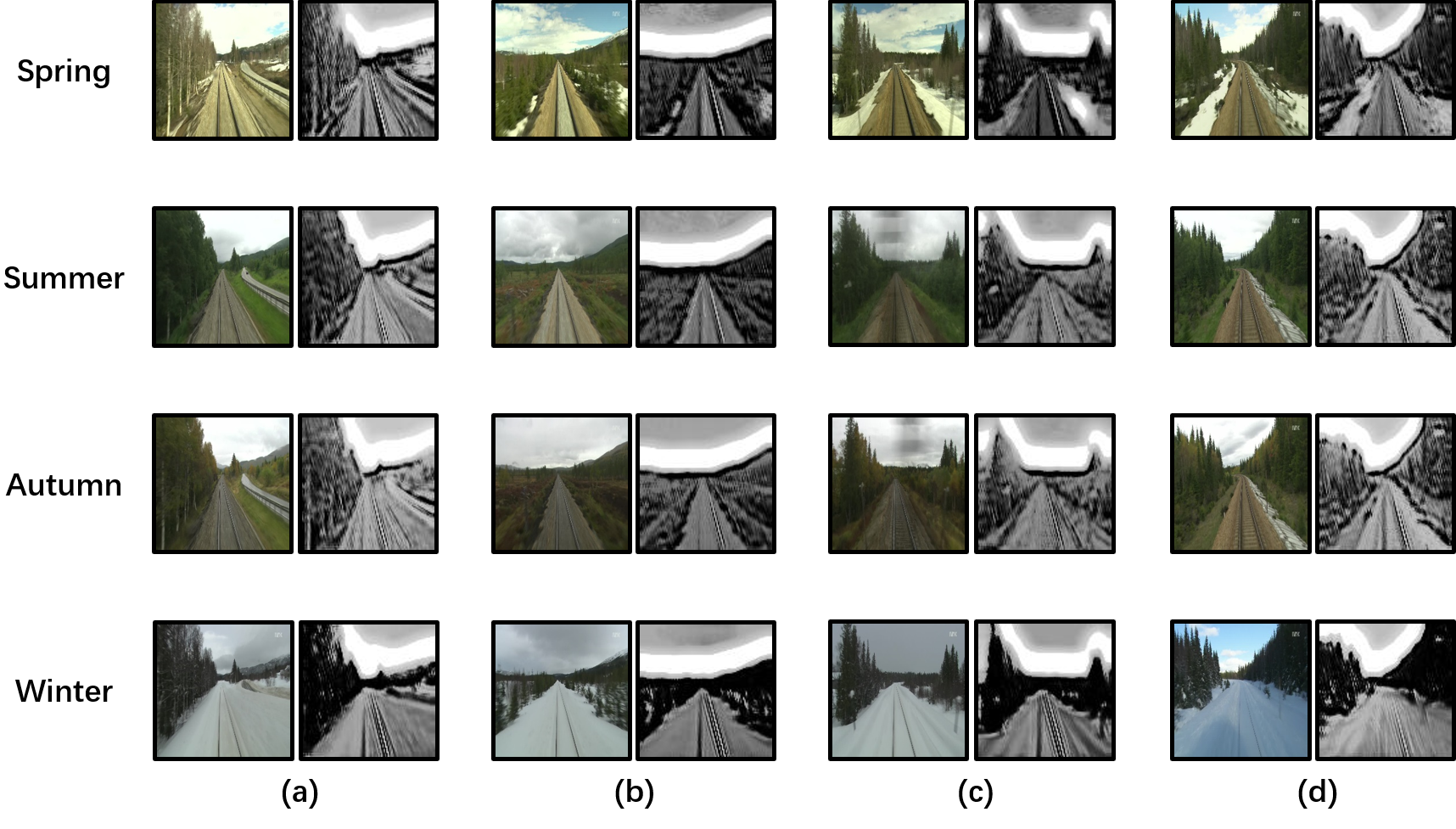}\\
  \caption{Some samples of various seasons and corresponding intrinsic representations in the testing set.
  The images from 1st to 4th rows belong to spring, summer, autumn, winter respectively.
  The original images (inputs of our intrinsic encoder) consist of images at the 4 seasons in the same place.
  The right images are corresponding intrinsic representations (outputs of our intrinsic encoder) of original images.
  The four sets of images (a), (b), (c) and (d)  are examples in different positions.
  }
  \label{results}
\end{figure}

%定性的 展示结果 %展示encoder的输出结果
In order to qualitatively demonstrate the effectiveness of proposed intrinsic encoder, some samples of various seasons and their intrinsic representations extracted by our intrinsic encoder are depicted in Fig.~\ref{results}.
Intrinsic representations are unsuited for direct visualization, because they only need to be robust to appearance changes and need not be similar to real images.
For the sake of observation, we implement histogram equalization to intrinsic representations in Fig.~\ref{results}.
The original images (inputs of our intrinsic encoder) consist of images at the 4 seasons (spring, summer, autumn, winter) in the same place.
The right images are corresponding intrinsic representations (outputs of our intrinsic encoder) of original images.
The four sets of images (a), (b), (c) and (d) are examples in different positions.

In our intrinsic space, road, tree, grass, mountain, skyline and sky are clearly distinguished.
Whether the green lush trees (in spring, summer ,and autumn) or bare trees (in winter), the trees can be detected and distinguished by our intrinsic encoder.
Regardless of which weather (for example, sunshine, cloudy ,and rainy) and category of cloud (for example, cumulus, cirrus, clear sky, stratus ,and cumulonimbus) the original images belong to, skyline and sky are marked consistently by our intrinsic encoder.
The qualitative results indicate that our intrinsic encoder can capture specific semantic contents (e.g., skyline, sky ,and tree) and removes the appearance changes caused by weathers and seasons.
As shown in Fig.~\ref{results} (c), this sample consists of two images (summer and autumn original images) which are disturbed by windshield wipers.
What is surprising is that our intrinsic encoder can suppress the noise automatically.

Different with semantic segmentation methods~\cite{semantic_seg,fcn} that train networks based on pixel-wise category labels with strong supervision,  our intrinsic encoder is trained based on weakly-supervised labels that only mark the seasons to which the unpaired training sets belong.
We employ proposed self-supervised encoder loss to train the intrinsic encoder capturing intrinsic representations of original images under various appearance changes, with no need for paired labeled training images corresponding to the same position under different conditions or pixel-wise category labels.
Note that we need not explicitly assign the specific semantic contents that should be distinguished.
Our intrinsic encoder can automatically capture condition-invariant intrinsic representations forced by proposed self-supervised encoder loss.
The intrinsic encoder implicitly captures some specific semantic contents, e.g., road, tree, grass, mountain, skyline ,and sky.

\begin{table*}[htbp]
  \centering
  \caption{
   Accuracy comparison with the baseline method SeqSLAM~\cite{seqslam} and the related state-of-the-art image-level place recognition method GANPL~\cite{gan-pl}.
   We report the place recognition accuracy with various sequence length $len = 1, 4$ in SeqSLAM~\cite{seqslam} and maximum distance $dis <= 1$.
  }
    \begin{tabular}{ccccccc}
    \toprule[1.25pt]
    \multirow{2}[0]{*}{\textbf{Methods}} & \multicolumn{2}{c}{\textbf{Spring-Winter}} & \multicolumn{2}{c}{\textbf{Summer-Winter}} & \multicolumn{2}{c}{\textbf{Autumn-Winter}} \\
    \cline{2-7}
    & 1     & 4         & 1     & 4        & 1     & 4     \\
    \toprule[1.25pt]
    SeqSLAM~\cite{seqslam} & 38.75 & 74.58  & 25.28 & 49.72  & 19.58 & 39.86  \\
    GANPL~\cite{gan-pl} & 61.39 & 89.03  & 53.47 & 84.72  & 43.60 & 74.86  \\
    Intrinsic Encoder & \textbf{81.67} & \textbf{94.72}  & \textbf{62.29} & \textbf{86.94}  & \textbf{44.58} & \textbf{76.37} \\
    \toprule[1.25pt]
    \end{tabular}%
  \label{acc}%
\end{table*}%

\subsubsection{Quantitative Results}
%定量的 place recognition
%acc
%precision-recall curve
%In order to quantitatively show the effect of our intrinsic encoder, we apply it to long-term place recognition task~\cite{seqslam}.
%Fig.~\ref{pr} depicts the precision-recall curves for various sequence length.
%Because our intrinsic encoder is the first method that can handle various environmental changes simultaneously without paired training data, we employ SeqSLAM~\cite{seqslam} (top) as a baseline for comparison.
%As the precision-recall curves show, our intrinsic representation (bottom) significantly boosts the performance of long-term place recognition.
%We locate between winter and other three seasons (spring, summer, autumn), and our intrinsic encoder achieves better results than SeqSLAM~\cite{seqslam} consistently.
%Consistent with the results in previous sequence-based works~\cite{seqslam,fast-seqslam} for long-term place recognition, we achieve higher precision and recall with larger sequence length.
%We almost achieves perfect performance with sequence length $len = 10$.
%Our intrinsic encoder implicitly captures specific semantic contents of original images under seasonal changes.
%As a result, intrinsic representation is condition-invariant of appearance changes and it describes the stable semantic structure of different places and scenes.

In order to quantitatively show the effect of our intrinsic encoder, we apply it to long-term place recognition task~\cite{seqslam}.
Because our intrinsic encoder is the first method that can handle various environmental changes simultaneously without paired training data, we employ SeqSLAM~\cite{seqslam} as a baseline for comparison.
In Table.~\ref{acc}, We compare our intrinsic encoder method with the baseline method SeqSLAM~\cite{seqslam} and the related state-of-the-art image-level place recognition method GANPL~\cite{gan-pl}.
Our intrinsic representation significantly boosts the performance of long-term place recognition.
We locate between winter and other three seasons (spring, summer, autumn), and our intrinsic encoder achieves better results than SeqSLAM~\cite{seqslam} consistently.
The proposed intrinsic encoder even boosts the accuracy more than $20\%$ than GANPL~\cite{gan-pl} with sequence length $len = 1$ for Spring-Winter.
It is worth noting that we generate only one intrinsic encoder rather than multiple generators in GANPL~\cite{gan-pl} for various seasons.
In testing stage, an intrinsic encoder can deal with several changing environmental conditions simultaneously.
However, a generator in GANPL~\cite{gan-pl} only aims at one changing environmental condition.

Consistent with the results in previous sequence-based works~\cite{seqslam,fast-seqslam} for long-term place recognition, we achieve higher accuracy with larger sequence length.
We still achieve remarkable performance without sequence-based information (sequence length $len = 1$) for challenging long-term place recognition.
Our intrinsic encoder implicitly captures specific semantic contents of original images under seasonal changes.
As a result, intrinsic representation is condition-invariant of appearance changes and it describes the stable semantic structure of different places and scenes.
Employing condition-invariant intrinsic representations, we significantly obtain higher location accuracy on long-term place recognition task.
The results demonstrate that our weakly-supervised intrinsic encoder can capture condition-invariant representations, which are effective and robust for long-term place recognition.
%并且不需要多种生成器in GANPL for various seasons，只需要唯一的encoder.

%%In order to clearly show the numerical compared results, the place recognition accuracy with various sequence length $len = 1, 4, 10$ and maximum distance $dis <= 1$ are given in Table.~\ref{acc}.
%In Table.~\ref{acc}, We compare our intrinsic encoder method with the related state-of-the-art image-level place recognition methods.
%Our intrinsic encoder even boosts the accuracy more than $20\%$ than GANPL~\cite{gan-pl} with sequence length $len = 1$ for Spring-Winter.
%The boost in performance is greater for smaller sequence length.
%%For every season, the place recognition performance improves significantly when we use proposed intrinsic representations to replace original images in SeqSLAM~\cite{seqslam}.
%We still achieve remarkable performance without sequence-based information (sequence length $len = 1$) for challenging long-term place recognition.
%Employing condition-invariant intrinsic representations, we significantly obtain higher location accuracy on long-term place recognition task.
%The results demonstrate that our weakly-supervised intrinsic encoder can capture condition-invariant representations, which are effective and robust for long-term place recognition.
%%并且不需要多种生成器in GANPL for various seasons，只需要唯一的encoder.
%It is worth noting that we generate only one intrinsic encoder rather than multiple generators in GANPL~\cite{gan-pl} for various seasons.

\section{Conclusion}
Motivated by the cognitive competence of human, we propose a novel intrinsic encoder to capture stable representations under drastic appearance changes, such as those caused by weathers, seasons, and day-night cycles in this paper.
The key characteristic of our intrinsic encoder is that it is trained based on weakly-supervised set-wise data rather than pixel-wise paired training data and one intrinsic encoder can solve various the environmental changes.
The proposed intrinsic encoder can implicitly capture specific semantic contents and suppress interference of appearance changes and noise.
Therefore, the captured image representations in intrinsic space are robust and adaptable for long-term environmental changes.

Experiments on Nordland dataset show that our image-level intrinsic encoder method boosts the performance on long-term place recognition task significantly.
The experimental results demonstrate that our weakly-supervised intrinsic encoder can capture the robust semantic representations and is effective for long-term place recognition.
It is worth noting that our weakly-supervised method is very suitable and effective for the situation where the paired training data are difficult to collect.
In practice, the collection of paired training data is laborious and time-consuming while the domain-based set-wise training data are relatively easier to collect.

{\small
\bibliographystyle{ieee}
\bibliography{egbib}
}

\end{document}